\documentclass{article}

\usepackage[final]{neurips_2021}
\usepackage{graphicx}
\usepackage{multicol}
\usepackage{hyperref}
\usepackage{comment}
\usepackage[normalem]{ulem}
\usepackage{color}
\usepackage{microtype}
\usepackage{amsmath}
\usepackage{ulem}
\usepackage{enumitem}
\usepackage{tablefootnote}
\usepackage[utf8]{inputenc} % allow utf-8 input
\usepackage[T1]{fontenc}    % use 8-bit T1 fonts
\usepackage{hyperref}       % hyperlinks
\usepackage{url}            % simple URL typesetting
\usepackage{booktabs}       % professional-quality tables
\usepackage{amsfonts}       % blackboard math symbols
\usepackage{nicefrac}       % compact symbols for 1/2, etc.
\usepackage{microtype}      % microtypography
\usepackage{xcolor}         % colors

\newcommand{\norm}[1]{\left\lVert#1\right\rVert_2 }
\def\A{\mathbf{A}}
\def\B{\mathbf{B}}
\def\C{\mathbf{C}}

\def \R{{\rm I\!R}}

\def\W{\mathbf{W}}
\DeclareMathOperator*{\argmin}{argmin}

\title{Kronecker Decomposition for GPT Compression}

\author{
  Ali Edalati$^2$, Marzieh Tahaei$^1$, Ahmad Rashid$^1$,\\ \textbf{ Vahid Partovi Nia$^1$, James J. Clark$^2$, Mehdi Rezagholizadeh$^1$ }\\
  $^1$ Huawei Noah Ark Lab\\
  $^2$ McGill University\\
  \texttt{ali.edalati@mail.macgill.ca} \\
}

\usepackage{ulem}
\begin{document}

\maketitle

\begin{abstract} 
GPT is an auto-regressive Transformer-based pre-trained language model which has attracted a lot of attention in the natural language processing (NLP) domain due to its state-of-the-art performance in several downstream tasks. The success of GPT is mostly attributed to its pre-training on huge amount of data and its large number of parameters (from ~100M to billions of parameters). Despite the superior performance of GPT (especially in few-shot or zero-shot setup), this overparameterized nature of GPT can be very prohibitive for deploying this model on devices with limited computational power or memory.    
This problem can be mitigated using model compression techniques; however, compressing GPT models has not been investigated much in the literature.
In this work, we use Kronecker decomposition to compress the linear mappings of the GPT-22 model. Our Kronecker GPT-2 model (KnGPT2) is initialized based on the Kronecker decomposed version of the GPT-2 model and then is undergone a very light pre-training on only a small portion of the training data with intermediate layer knowledge distillation (ILKD). Finally, our KnGPT2 is fine-tuned on down-stream tasks using ILKD as well. 
We evaluate our model on both language modeling and General Language Understanding Evaluation  benchmark tasks and show that with more efficient pre-training and similar number of parameters, our KnGPT2 outperforms the existing DistilGPT2 model significantly.

\end{abstract}

\section{Introduction}
\label{sec:1}

Recently, development and deployment of pre-trained language models (PLMs) has improved the performance of NLP models significantly ~\citep{devlin2018bert,radford2019language,yang2019xlnet,shoeybi2019megatron,radford2019language}. PLMs are mostly Transformer-based models, which are pre-trained on enormous unlabeled data. 
Although Transformer-based PLMs are powerful in performance, their huge size is a barrier for efficient training or inference of these models on lower capacity devices with memory, computation and energy constraints. Therefore, there has been a growing volume of literature focused on developing frameworks for compressing these large PLMs.
Like other deep learning models, the main directions of model compression for PLMs are using following methods in isolation or combination: low-bit quantization \citep{gong2014compressing,prato2019fully}, pruning \citep{han2015deep}, knowledge distillation (KD) \citep{hinton2015distilling} and matrix decomposition (\cite{yu2017compressing,lioutas2020improving}).

PLMs can be devided into encoder-based and auto-regressive models such as the BERT~\citep{devlin2018bert, liu2019roberta} and GPT~\citep{brown2020language} family respectively. Although the size of BERT family models is usually smaller than the GPT family, compressing the BERT family has been investigated much more in the literature (e.g. DistilBERT~\citep{sanh2019distilbert}, TinyBERT~\citep{jiao2019tinybert}, MobileBERT~\citep{sun2020mobilebert}, ALP-KD~\citep{passban2020alp}, MATE-KD~\citep{rashid2021mate}, Annealing-KD~\citep{jafari2021annealing} and BERTQuant~\citep{zhang2020ternarybert}). On the other hand, to the best of our knowledge, the GPT family has barely a handful of compressed models, among them the DistilGPT2\footnote{https://transformer.huggingface.co/model/distil-gpt2} model is very prominent. The DistilGPT2 model is heavily pre-trained for 3 epochs on the large OpenWebText dataset\footnote{https://huggingface.co/datasets/openwebtext}. Moreover, it is evident in the literature that the GPT model cannot compete with BERT on natural language understanding (NLU) tasks~\citep{liu2021gpt}. Therefore, developing an efficient compressed GPT model with comparable NLU performance is still an open problem.         

In this paper, we use Kronecker decomposition, which has been recently  used for BERT compression \citep{tahaei2021kroneckerbert}, for compression of the GPT-2 model (we refer to our model as KnGPT2 in this paper).  We use Kronecker decomposition to represent the weight matrices of linear layers in GPT-2 by smaller matrices which can reduce the size and computation overhead. 
We use Kronecker decomposition to compress the embedding and Transformer layers of GPT-2. For Transformer layers, the linear layers of multi-head attention (MHA) and the feed-forward network (FFN) blocks of Transformer layers are decomposed into Kronecker layers. 

 %There are some previous works \cite{khrulkov2019tensorized, li2018slim} that tried to reduce memory consumption of embedding layers and our method proposes an efficient methods for compressing the embedding layer's wight matrices.

Kronecker decomposition leads to reduction in expressiveness of the model. We use a very light pre-training with intermediate layer knowledge distillation (ILKD) to address this issue, which improves the performance of the compressed model significantly. It is worth mentioning that for our pre-training, we use $1/10^{\text{th}}$ of the DistilGPT2's pre-training data (i.e. OpenWebText) only for 1 epoch (instead of 3 epochs in DistilGPT2). 
Furthermore, in this paper, our framework is applied to GPT-2 but it can be easily exploited to compress other models as well. To summarize contributions of this paper, we mention the following points:
\begin{itemize}
    \item To the best of our knowledge, we are the first work which  uses Kronecker decomposition for compression of the GPT model.
%    \item Implementing the kronecker decomposition on different versions of GPT2 as a base model with and without KD.
    %\item Investigating the capacity gap by applying KD with different size teachers
    \item Our KnGPT2 model improves training efficiency and performance of the DistilGPT2 model significantly. 
    \item We evaluate the performance of our KnGPT2 on both language modeling and the GLUE benchmark tasks. 
\end{itemize}

\section{Related Works}
\label{sec:2}
In this section, first, we review some of previous works that have deployed Kronecker decomposition for compression of deep learning models. Then, some works related to GPT compression are covered.
%\subsection{Kronecker Factorization}
\cite{zhou2015compression} is the first work that used summation of multiple Kronecker products to compress the weight matrices in fully-connected networks and small convolutional neural networks. \cite{thakker2019compressing} proposed a hybrid method which separates the weight matrices into an upper and a lower part, upper part remains untouched but the lower part decomposes to Kronecker products. They used this approach for small language models to be utilized on internet of things (IoT) applications. Recently, \cite{thakker2020compressing} extended the mentioned hybrid method to non-IoT applications by adding a sparse matrix to the Kronecker products.
\cite{tahaei2021kroneckerbert} has deployed a similar approach to ours to compress BERT which achieved promising results but to the best of our known, this work is the first attempt for GPT compression using kronecker decomposition.

%\subsection{GPT Compression}
DistilGPT2 \footnote{For further details, see https://huggingface.co/distilgpt2 } is one of the most successful and well-known compressed versions of GPT-2 which is considered as a baseline in this paper. DistilGPT2 has 82M parameteres compared to 124M parameters for $\text{GPT-2}_{\text{Small}}$ and is trained using KD on OpenWebTextCorpus which is a reproduction of OpenAI's WebText dataset.

\section{Methodology}
In this section, we first provide some background on Kronecker product and its mathematical properties. We then explain how Kronecker factorization can be used for the compression of linear layers and subsequently for the GPT model.

\subsection{Kronecker Product}
The Kronecker product is a matrix operation (denoted by $\otimes$) which takes two matrices as input and generates a block matrix as output. Assume that $\A$ is a matrix $\in \R^{m_1\times n_1}$ and $\B$ is a matrix $\in \R^{m_2\times n_2}$, $\A \otimes \B$ is equal to a block matrix $\in \R^{m \times n}$, where $m=m_1m_2$, $n=n_1n_2$ and each block $(i,j)$ is obtained by multiplying element $a_{ij}$ by matrix $\B$. %Kronecker product of two matrices is defined as:
\begin{equation}
    \mathbf{A}\otimes\mathbf{B} = \begin{bmatrix}
  a_{11} \mathbf{B} & \cdots & a_{1n}\mathbf{B} \\
             \vdots & \ddots &           \vdots \\
  a_{m1} \mathbf{B} & \cdots & a_{mn} \mathbf{B},
\end{bmatrix}
\end{equation}

Kronecker product  has attractive abstract algebraic properties such as 
$$
\A \otimes (\B+\C) = \A \otimes \B + \A \otimes \C,\quad (\A\otimes \B)^{-1} = \A^{-1} \otimes \B^{-1}, \quad (\A\otimes \B)^\top = \A^\top \otimes \B^\top,
$$
for more details see  \cite{henderson1983history}.
The interesting properties of the Kronecker product makes it an attractive tool for decomposition of large matrices. %In is common to use Kronecker in product to express large systems in physics, for instance the total Hamiltonian of a system can be written as the Kronecker product of each individual system. 
The Kronecker product is also a flexible method to simplify the notation of large block matrices, both in  linear mixed effect models and multilevel models \cite{goldstein2011multilevel}. It is also a well-known technique to represent large repetitive structured graphs using the Kronecker product \cite{leskovec2010kronecker}. One of the most important characteristics of a matrix is its determinant and it is well-known that for two square matrices $\A$ and $\B$ of size $n$, and $m$, $|\A\otimes \B| =|\A|^n |\B|^m $. This property explains the superiority of Kronecker compared to the other decomposition methods for large matrices. By choosing the right $n$ and $m$, a large matrix $\W=\A\otimes \B$ can be decomposed to much smaller matrices such that the above determinant equation holds.

 %Figure \ref{fig:1} illustrates kronecker product of two small matrices. 

\subsection{GPT-2 Compression using Kronecker Factorization}
We can represent a weight matrix, $\W\in \R^{m\times n}$, by two smaller matrices, $\A\in \R^{m_1\times n_1}$ and $\B \in \R^{m_2\times n_2}$ such that $\W=\A\otimes \B$ and $m=m_1m_2$, $n=n_1n_2$. This leads to reduction in the number of parameters from $mn$ for the original matrix to $m_1n_1+m_2n_2$ for the Kronecker factorized version.
%\sout{\subsubsection{Memory reduction}
%Kronecker factorization enables us to keep two smaller matrices, which have $m_1n_1+m_2n_2$ parameters, instead of keeping the bigger weight matrix which has $mn$ parameters.
For example, assume that size of $\W$ is $1024\times 1024$, we can represent it by two matrices of sizes $512\times 512$ and $2\times 2$ for which the compression factor will be roughly equal to 4.
%\begin{equation}
%\label{eq:1}
%\centering
%    ActivationFunction(Wx+bias) = y
%\end{equation}
%Equation \ref{eq:1} shows how output of a linear layer, y, is calculated given the input, x. If we represent weight matrix, $W$ by two matrices $A$ and $B$  using kronecker factorization, Equation \ref{eq:1} changes to Equation \ref{eq:2}. 
%\begin{equation}
%\label{eq:2}
%\centering
%    ActivationFunction((A\otimes B)x+bias) = y
%\end{equation}
%We can simply reconstruct the $\W$ by obtaining $\A \otimes \B$ and multiply it by $\x$ to compute output of the Kronecker layer.

In the following we explain how this work uses Kronecker factorization for compression of the GPT-2 model. In large language models, embedding layer usually takes a large portion of the memory. Let $ \W^E \in \R^{v\times d}$ be the lookup table for the input embedding  where $v$ is the vocabulary size and $d$ is the embedding dimension. To compress the embedding layer using Kronecker decomposition we use the same method as in \cite{tahaei2021kroneckerbert}. We define $\A^E \in \R^{v\times d/f}$ and $\B^E \in \R^{1\times f}$, where $f$ is a factor of $d$. There are two reasons for this decision: first, similar to $\W^E$, in the $\A^E$ matrix every row will indicate embedding of a single word. Second, the embedding of each word, $E_i$, can be obtained by $\A^{E}_i \otimes \B$ therefore the computation complexity of this operation is $\mathcal O(d)$ which is very efficient. 

The transformer architecture is composed of $N$ identical layers each having MHA followed by FFN.
%\sout{MHA and FFN are the main blocks that compose the transformer part of GPT2.}
In the MHA module, there are linear layers which calculate the Query, Key and Value by multiplying the input vector by $\W^Q, \W^K,\W^V$, respectively. Also, in the FFN module, there are two fully connected layers that can be represented as $\W^{c_\mathrm{fc}} and \W^{c_{\mathrm{proj}}}$. In this work, all of the mentioned weight matrices at different heads and layers of the transformer are decomposed into Kronecker factors.

For initialization, similar to \cite{tahaei2021kroneckerbert}, the Kronecker factors $\hat\A$ and $\hat\B$ are estimated from the corresponding weight matrix $\W$ in the original uncompressed pre-trained  model using the solution to the nearest Kronecker problem 
$$(\hat\A, \hat\B) = \argmin\limits_{(\A,\B)} \norm{\mathrm \W-\A\otimes \B}^2.$$ The solution to this optimization can be found by rank-1 singular value decomposition (SVD) approximation of the reshaped $\W$, see \cite{van2000ubiquitous} for details.
\subsection{Knowledge Distillation}
\label{sec:3.5.1}
In this section, the knowledge distillation method used for training the KnGPT2 model is explained.  The same method is used in the pre-training and fine-tuning stages. %\subsubsection{Intermediate knowledge distillation}

Let $T$ and $S$ represent the teacher model, GPT-2,  and the student model, KnGPT2, respectively. For a batch of data $(\mathbf x, \mathbf y)$, $E^S$ and $E^T$ are outputs of the embedding layers of the student and teacher models respectively. Also, $\text{Att}^{S}_l$ and $\text{Att}^{T}_l$ are the attention distributions obtained by applying softmax on the scaled dot product between query and key. $H^{S}_l$ and $H^{T}_l$ are the hidden state outputs of the layer $l$. Note that by using the Kronecker factorization, like other decomposition methods, the number of layers and dimensions of the output matrices in the student model remain intact so we can directly obtain the difference of output of a specific layer in student an teacher model without the need for projection. 
For the embedding layer we use the mean squared error (MSE) between the teacher's and student's embeddings:
\begin{equation}
\label{eq:6}
    \centering
    L_\text{Embedding}(x)=\text{MSE}\{E^{S}(x), E^{T}(x)\}
\end{equation}
For the MHA modules, similar to \cite{wang2020minilm}, we use Kullback–Leibler divergence (KL) between the attention distributions of the student and the teacher.
\begin{equation}
\label{eq:7}
    \centering
    L_\text{Attention}(x)=\sum_{l} \text{KL}\{\text{Att}^{S}_{l}(x), \text{Att}^{T}_{l}(x)\}
\end{equation}
For the FFN modules, we simply use the MSE between the output of the second fully connected layer in the student and teacher:
\begin{equation}
\label{eq:8}
    \centering
    L_\text{Hidden States}(x)=\sum_{l} \text{MSE}\{H^{S}_{l}(x), H^{T}_{l}(x)\}
\end{equation}
The final loss is calculated a linear combination of the above losses as well as the cross entropy loss. 
\begin{equation}
\label{eq:9}
    \centering
    \text{Loss}(x,y)=\sum _{(x,y)} \alpha_1L_\text{Embedding}(x)+\alpha_2L_\text{Attention}(x)+\alpha_3L_\text{Hidden States}(x)+\alpha_4L_\text{Cross Entropy} (x,y)
\end{equation}

After decomposing the teacher model, GPT-2, into KnGPT2, the performance of the model drops significantly. This drop is mainly because of the approximation of  linear weight matrices using the  corresponding Kronecker factors. Therefore, pre-training of the compressed model on a small corpus for a few epochs  is necessary to retrieve the information which are lost during decomposition. Inspired by \cite{jiao2019tinybert}, we pre-trained the model on a small portion, 10\%, of the OpenWebText dataset \cite{Gokaslan2019OpenWeb} for one epoch and we used the KD method which is discussed in Section \ref{sec:3.5.1} to improve the performance of the compressed model. %Table \ref{tab:3} and Table \ref{tab:6} show the effect of kronecker decomposition on perplexity of compressed model and how pre-training has improved the performance.

%\subsubsection{Model configuration}
%In this work, our baseline is DistilGPT2 which has roughly 80 million parameters so we should compress the GPT model such that the kronecker model have approximately the same number of parameters with the baseline to have a fare comparison. We compressed GPT2 (12 layers) and GPT-medium (24 layers) and since they have different configuration, we compressed them with different settings. Note that for compressing GPT2, compressing only half of the transformer layers was enough to reach the desired number of parameters. 
\begin{table}
  \centering
  \begin{tabular}{lccc}
    \toprule
    Model      & Embedding & Q,K,V & $\mathrm{FFN}^*$ \\
    \midrule
    $\text{GPT-2}_{\text{Small}}$      &  $50527\times 768$ & $768\times 768$ & $3072\times 768$ \\ 
    DistilGPT2 &  $50527\times 768$ & $768\times 768$ & $3072\times 768$ \\ 
    KnGPT2 &$A:50527\times 384$, $B:1\times 2$ & $A:384\times 768$, B:$2\times 1$ & $A:1536\times 768$, $B:2\times 1$
    \\
    \bottomrule
  \end{tabular}
  \caption{This table shows configuration of the models. Note that FFN block has two projections that shape of one is the transpose of the other one and here, only shape of one of them is mentioned. Also, for KnGPT2, mentioned shapes for transformer layer belong to half of the layers that are decomposed -layers with odd numbers- and shape of the other half are the same with the GPT-2 model.}
  \label{tab:1}
\end{table}

\begin{table}
  \centering
  \begin{tabular}{lcccccc}
    \toprule
    Phase     &  Batch size    &  Learning rate
    & $\alpha_1$ & $\alpha_2$ & $\alpha_3$ & $\alpha_4$\\
    \toprule
    Pre-training & 1  & 0.00025 & 0.5 & 0.5 & 0.5 & 0.1   \\
    Fine-tuning     & 16 & 2e-5 & 0.5 & 0.5 & 0.5 & 0.02\\
    \bottomrule
  \end{tabular}
  \caption{hyper-parameters that are used for pre-training and fine-tuning.}
  \label{tab:2}
\end{table}

\section{Experiments}

We evaluated our proposed algorithm, KnGPT2, on language modeling and text classification. For language modeling we use the Wikitext-103~\cite{meritypointer} dataset.For classfication we use  seven of the classification tasks of the General Language Understanding Evaluation (GLUE) benchmark~\citep{wang2019glue}. These datasets can be broadly divided into 3 families of problems. Single set tasks which include linguistic acceptability (CoLA) and sentiment analysis (SST-2), similarity and paraphrasing tasks (MRPC and QQP), and  inference tasks which include Natural Language Inference (MNLI and RTE) and Question Answering (QNLI). 

\subsection{Experimental Setup}
The KnGPT2 model is compressed from the $\text{GPT-2}_{\text{Small}}$~\cite{radford2019language} model. $\text{GPT-2}_{\text{Small}}$ is 124 million parameters. Our baseline is DistilGPT2 which has about 82 million parameters so our KnGPT2 model is compressed to the same size (83 million parameters) for a fair comparison. To achieve this, we compress half the layers of transformer block (odd numbered ones) in addition to the embedding layer by a factor of 2. Table \ref{tab:1} shows the configuration of the models. Table \ref{tab:2} shows hyper-parameters that are used for pre-training and fine-tuning.
% \subsection{Baseline}
% We used DistilGPT2\footnote{For further details, see \hyperref[https://huggingface.co/distilgpt2]{https://huggingface.co/distilgpt2}} as the baseline model which has roughly 82 million parameters and is pre-trained on the OpenWebText dataset. To measure its perplexity, the model is trained on Wikitext103 \cite{meritypointer}. Also, DistilGPT2 is fine-tuned (first without KD and then using vanilla KD  while the teacher was GPT2) on GLUE\footnote{For further details, see \hyperref[https://gluebenchmark.com/]{https://gluebenchmark.com/}} tasks \cite{wang2019glue} to be compared to out model.

\subsection{Pre-training}
After the base model is compressed using Kronecker decomposition, performance of the compressed model drops significantly since the weight matrices with the Kronecker factors are approximate. Pre-training on a relatively small data set for one epoch helps in retrieving the accuracy. Therefore, KnGPT2 is pre-trained on 10\% of OpenWebText which is 10 times less the DistiGPT2 model. As shown on Table~\ref{tab:3} the training time for KnGPT2 is much faster as well. 
\begin{table}
  \centering
  \begin{tabular}{lccc}
    \toprule
         & $\text{GPT-2}_{\text{Small}}$ & DistilGPT2 & KnGPT2 \\
    \toprule
    % Perplexity & 18.8 & 23.7 & 20.5\\
    $\mathrm{Parameters}^*$ & 124 & 82 & 83 \\
    Training time (hrs) & - & >90\tablefootnote{This number is presented in \cite{sanh2019distilbert} for training DistilBERT by the same authors. That uses the same KD algorithm and dataset for pre-training but is applied to BERT rather than GPT. Using a similar hardware we expect this number to be larger for DistilGPT} & 6.5 \\
    Dataset size (GB) & 40 & 38 & 3.2\\
    \bottomrule
  \end{tabular}
  \caption{Training details for GPT-2 compression. Note that number of parameters of the models are reported excluding the output embedding layer in language modelling which is not compressed, is equal to row $\mathrm{Parameters}^*$ }
  \label{tab:3}
\end{table}

\subsection{Results}

We measure the performance of our compressed model on two tasks. First we evaluate on language modeling using the Wikitext-103 dataset. The initialized models a are first trained on this dataset and then evaluated on the provided test set. The results are shown on Table~\ref{tab:lm}. Although the DistilGPT2 is pre-trained longer and on a larger dataset the KnGPT2 achieves a lower perplexity. 

\begin{table}[t]
\centering
%\fontsize{10}{12}\selectfont
\begin{tabular}{lccc}
\toprule
& $\text{GPT-2}_{\text{Small}}$ & DistilGPT2 & KnGPT2 \\
\midrule
Perplexity & 18.8 & 23.7 & 20.5 \\
\bottomrule
\end{tabular}
\caption{Test Perplexity on WikiText-103.}
\label{tab:lm}
\end{table}

Next the performance of the models is evaluated on  both the development (Table \ref{tab:4}) and test (Table \ref{tab:5}) sets of seven datasets of the GLUE benchmark. In addition to employing the cross-entropy loss for fitting the labels we also experiment with KD. For DistilGPT, we apply the basic KD algorithm also referred to in the literature as Vanilla KD~\cite{jafari2021annealing}. For KnGPT2 we apply intermediate layer distillation as well as Vanilla KD. For DistilGPT since the number of layers between the teacher and the student are different, it is not clear which teacher layer should be distilled to which student layer. Although there has been work on intermediate distillation for mismatched layers for BERT~\cite{passban2020alp}, extensive experimentation is required to conclude the best practivce for GPT. 

On the dev set results (Table~\ref{tab:4}), we observe that KnGPT2 performs better than DistilGPT2 for most datasets and on average. If we apply KD we observe that it is better on all datasets compared to DistilGPT2. Another interesting result is that Vanilla KD does not improve DistilGPT2 fine-tuning. The test set results on Table~\ref{tab:5} follow the same trend as the dev results. Interestingly KnGPT2 with KD reaches close to the $\text{GPT-2}_{\text{Small}}$ performance on average.

\begin{table}
 
  \centering
  \begin{tabular}{l|cccccccc}
    \toprule
      Model   & CoLA & RTE & MRPC & SST-2 & MNLI & QNLI & QQP & Average \\
    \toprule
    $\text{GPT-2}_{\text{Small}}$ & 47.6 & 69.31 & 87.47 & 92.08 & 83.12	& 88.87 & 90.25 & 79.81\\
    \midrule
    DistilGPT2  & 38.7 & 65.0 & 87.7 & 91.3 & 79.9 & 85.7 & 89.3 & 76.8 \\
    DistilGPT2 + KD & 38.64 & 64.98 & 87.31	 & 89.80 & 80.42 & 86.36 & 89.61 & 76.73 \\
    KnGPT2  & 37.51 & \textbf{70.4} & \textbf{88.55} & 88.64 & 78.93 & 86.10 & 88.87 & 77 \\
    KnGPT2 + ILKD & \textbf{45.36} & 69.67 & 87.41 & \textbf{91.28}	& \textbf{82.15} & \textbf{88.58} & \textbf{90.34} & \textbf{79.25} \\
    \bottomrule
  \end{tabular}
  \caption{This table shows performance of the models on dev set of GLUE tasks. Note that $\text{GPT-2}_{\text{Small}}$ is used as teacher for KD.}
  \label{tab:4}
\end{table}

% \begin{table}
%  \label{tab:4}
%   \centering
%   \begin{tabular}{l|c|cccccccc}
%     \toprule
%       Model  & KD & CoLA & RTE & MRPC & SST-2 & MNLI & QNLI & QQP & Average \\
%     \toprule
%     GPT2 & - & 47.6 & 69.31 & 87.47 & 92.08 & 83.12	& 88.87 & 90.25 & 79.81\\
%     \midrule
%     DistilGPT2 & - & 38.7 & 65.0 & 87.7 & 91.3 & 79.9 & 85.7 & 89.3 & 76.8 \\
%     DistilGPT2 & vanila & 38.64 & 64.98 & 87.31	 & 89.80 & 80.42 & 86.36 & 89.61 & 76.73 \\
%     KnGPT2 & - & 37.51 & \textbf{70.4} & \textbf{88.55} & 88.64 & 78.93 & 86.10 & 88.87 & 77 \\
%     KnGPT2 & intermediate & \textbf{45.36} & 69.67 & 87.41 & \textbf{91.28}	& \textbf{82.15} & \textbf{88.58} & \textbf{90.34} & \textbf{79.25} \\
%     \bottomrule
%   \end{tabular}
%   \caption{This table shows performance of the models on dev set of GLUE tasks. Note that GPT2 is used as teacher for KD.}
% \end{table}

\begin{table}
 
  \centering
  \begin{tabular}{l|cccccccc}
    \toprule
      Model & CoLA & RTE & MRPC & SST-2 & MNLI & QNLI & QQP & Average \\
    \toprule
    $\text{GPT-2}_{\text{Small}}$  & 44.0 & 63.2 & 84.5 & 92.8 & 81.75 & 88.7 & 88.0 & 77.56 \\
    \midrule
    DistilGPT2  & 32.4 & 61.9 & 84.3 & 90.8 & 79.55 & 85.4 & 87.3 & 74.52 \\
    DistilGPT2 + KD & 33 & 61.5 & 84.4 & 90.7 & 79.85 & 85.7 & 87.6	& 74.67 \\
    KnGPT2  & 36.7 & \textbf{64.4} & 84.5 & 89.0 & 78.45 & 85.6 & 86.5 & 75.02 \\
    KnGPT2 + ILKD  & \textbf{41.8} & 63.7 & \textbf{86.5} & \textbf{91.5} & \textbf{81.6} & \textbf{88.4} & \textbf{88.5} & \textbf{77.42} \\
    \bottomrule
  \end{tabular}
  \caption{This table shows performance of the models on test set of GLUE tasks. Note that $\text{GPT-2}_{\text{Small}}$ is used as teacher for KD.}
  \label{tab:5}
\end{table}

% \begin{table}
%  \label{tab:5}
%   \centering
%   \begin{tabular}{l|c|cccccccc}
%     \toprule
%       Model & KD  & CoLA & RTE & MRPC & SST-2 & MNLI & QNLI & QQP & Average \\
%     \toprule
%     GPT2 & - & 44.0 & 63.2 & 84.5 & 92.8 & 81.75 & 88.7 & 88.0 & 77.56 \\
%     \midrule
%     DistilGPT2 & - & 32.4 & 61.9 & 84.3 & 90.8 & 79.55 & 85.4 & 87.3 & 74.52 \\
%     DistilGPT2 & vanila & 33 & 61.5 & 84.4 & 90.7 & 79.85 & 85.7 & 87.6	& 74.67 \\
%     KnGPT2 & - & 36.7 & \textbf{64.4} & 84.5 & 89.0 & 78.45 & 85.6 & 86.5 & 75.02 \\
%     KnGPT2 & intermediate  & \textbf{41.8} & 63.7 & \textbf{86.5} & \textbf{91.5} & \textbf{81.6} & \textbf{88.4} & \textbf{88.5} & \textbf{77.42} \\
%     \bottomrule
%   \end{tabular}
%   \caption{This table shows performance of the models on test set of GLUE tasks. Note that GPT2 is used as teacher for KD.}
% \end{table}

\subsection{Ablation Study}
We performed an experiment to study the effect of KD on the pre-training of KnGPT2. In this experiment we used Wikitext-103 as our pre-training dataset. We compare four models and evaluate on LM  as well as on classification using the MNLI dataset from GLUE. As shown on Table~\ref{tab:6} we compare KnGPT2 without pre-training, with language modeling pre-training only, with KD pre-training only and both language modeling and KD pre-training. Note that we apply ILKD, discussed before for finetuning, as our KD algorithm. We observe that pre-training is important for good performance on the downstream task but lower perplexity on LM is not always a good indicator of better downstream performance. 

\begin{table}[h]
\centering
%\fontsize{10}{12}\selectfont
\begin{tabular}{lccc}
\toprule
Model & Wikitext-103(pp.) & MNLI (f1)\\
\midrule
KnGPT2 & 28608 & 69.33\\
KnGPT2 + LM & \textbf{21.94} & 77.87 \\
KnGPT2 + KD & 144.1 & 77.50 \\
KnGPT2 + LM + KD & 23.04 & \textbf{77.97} \\
\bottomrule
\end{tabular}
\caption{Ablation on the effect of pre-training with KD on language model and MNLI classification}
\label{tab:6}
\end{table}

% \begin{table}
%   \centering
%   \begin{tabular}{lccccc}
%     \toprule
%      & $\mathrm{GPT2}$ & $\mathrm{KnGPT2}^{0}$ 
%     & $\mathrm{KnGPT2}^{1}$ & $\mathrm{KnGPT2}^{2}$
%     & $\mathrm{KnGPT2}^{3}$\\
%     \toprule
%     Perplexity & 18.8 & 28608 & 21.94 & 144.1 & 23.04 \\
%     SST-2 accuracy & 92.08 & 81.42 & 87.5 & 87.5 & 87.72\\
%     MNLI accuracy & 83.12 & 69.33 & 77.87 & 77.50 & 77.97 \\
%     \bottomrule
%   \end{tabular}
%   \caption{This table shows performance of the models on Wikitext-103 where superscript of the name of the models refer to the used pre-training method.}
%   \label{tab:6}
% \end{table}
% \subsubsection{KD at fine-tuning}
% To examine the effect of intermediate KD on performance of the model, KnGPT2 is fine-tuned with and without the intermediate KD method. Table \ref{tab:4} and Table \ref{tab:5} shows that using intermediate KD has improved the performance significantly.

\section{Conclusion}
In this paper, we compressed GPT-2 by compressing linear layers of a GPT model using Kronecker decomposition. Our model is pre-trained on a relatively small (10 times smaller than the dataset used for baseline) dataset which makes the pre-training much faster. Our proposed model significantly outperformed the baseline on the GLUE benchmark. Using KD can help to further reduce the performance drop of the compressed model. Using Kronecker decomposition on larger GPT models and for higher compression factors are two interesting future directions.

\bibliography{references}
\bibliographystyle{acl_natbib}
% \bibliography{natbib}
% \bibliographystyle{abbrv}

\end{document}